\pdfoutput=1

\documentclass[11pt]{article}
\usepackage{multirow}
\usepackage{ACL2023}

\usepackage{times}
\usepackage{latexsym}

\usepackage[T1]{fontenc}

\usepackage[utf8]{inputenc}

\usepackage{tablefootnote}

\usepackage{pdfpages}

\usepackage{graphicx}

\usepackage{microtype}

\usepackage{inconsolata}

\title{Towards Mitigating Perceived Unfairness in Contracts from a Non-Legal Stakeholder’s Perspective}

\author{Anmol Singhal, Preethu Rose Anish, Shirish Karande, and Smita Ghaisas
\\ TCS Research, India \\
\texttt{\{anmol.singhal, preethu.rose, shirish.karande, smita.ghaisas\}@tcs.com}}

\begin{document}
\maketitle
\begin{abstract}
Commercial contracts are known to be a valuable source for deriving project-specific requirements. However, contract negotiations mainly occur among the legal counsel of the parties involved. The participation of non-legal stakeholders, including requirement analysts, engineers, and solution architects, whose primary responsibility lies in ensuring the seamless implementation of contractual terms, is often indirect and inadequate. Consequently, a significant number of sentences in contractual clauses, though legally accurate, can appear unfair from an implementation perspective to non-legal stakeholders. This perception poses a problem since requirements indicated in the clauses are obligatory and can involve punitive measures and penalties if not implemented as committed in the contract. Therefore, the identification of potentially unfair clauses in contracts becomes crucial. In this work, we conduct an empirical study to analyze the perspectives of different stakeholders regarding contractual fairness. We then investigate the ability of Pre-trained Language Models (PLMs) to identify unfairness in contractual sentences by comparing chain of thought prompting and semi-supervised fine-tuning approaches. Using BERT-based fine-tuning, we achieved an accuracy of 84\% on a dataset consisting of proprietary contracts. It outperformed chain of thought prompting using Vicuna-13B by a margin of 9\%. 
\end{abstract}

\section{Introduction}

Contracts represent legally binding agreements among all participating stakeholders. Contract drafts typically undergo several iterations before they can be finalized, owing to their business criticality. The iterations involve extensive negotiations among legal professionals on accepting or rejecting parts of drafts put forth by other parties. This collaborative exchange of perspectives is critical in ensuring that all involved parties ultimately view the resulting contracts as acceptable.

Commercial contracts between enterprises are invaluable sources for deriving project-specific requirements \cite{1}. The requirements indicated in the contractual clauses are obligatory. Therefore, it is important that they are perceived as practical and implementable by stakeholders responsible for meeting those commitments. However, these non-legal stakeholders, such as project managers, requirement analysts, engineers, solution architects, and delivery personnel, are not directly involved in contract negotiations. While in-house legal experts do consult various teams to understand the feasibility and scope of individual projects before drafting a contract, there is considerable room for oversight in the high-level discussions that take place \cite{contract-neg}. The issue is further aggravated because the non-legal stakeholders share a limited understanding of Legalese, the specialized language used in contract writing. Consequently, many clauses can seem unfair to non-legal stakeholders despite being the results of iterative negotiations by legal experts. The perception of all involved stakeholders needs to be adequately captured during contract drafting for smooth governance \cite{Karabulut2020TheRA}. 

In this work, we examine contracts from an angle of fairness, or lack thereof, from non-legal stakeholders' perspectives who must implement the obligatory requirements outlined within contracts. To achieve this goal, we conducted an empirical study involving legal and non-legal stakeholders who are often participants in contract negotiations and/or responsible for implementing the projects according to contract obligations. Our primary objective was to gain insights into their perceptions regarding fairness within contracts. Our analysis revealed multiple scenarios of unfairness that can occur within contract text. Some clauses are interpreted as ‘clearly unfair’ by both legal and non-legal stakeholders because they seem to be causing an obvious imbalance in parties’ rights or obligations. Such clear unfairness is often encountered in clauses that address critical project management aspects, including unilateral termination, unilateral modification of terms, restrictions on indemnity claims, and jurisdictional matters. Our study also revealed that some clauses containing requirements may be considered ‘potentially unfair’ by non-legal stakeholders because of a risk of non-compliance arising due to the inclusion of ambiguous language. Such clauses can make certain parties vulnerable by significantly impacting the cost and effort associated with fulfilling the obligatory requirements. For instance, the sentence \textit{The vendor should implement appropriate measures to ensure a level of security commensurate to the risk} vaguely specifies the measures required to implement the security controls. If the expectation regarding what constitutes `appropriate' security controls differs between the vendor and the customer, the customer may perceive the vendor's enforcement of security controls as unfair and a breach of contract. 



Based on the findings of the empirical study, we envisage building an automated assistant tailored to streamline the process of contract review by bringing in the perspectives of non-legal stakeholders. The assistant would: (1) identify contractual sentences likely to be perceived as unfair by non-legal stakeholders; (2) distinguish between clearly unfair and potentially unfair clauses; (3) recommend questions to be asked to the parties involved, thereby assisting in the resolution of the unfairness within the sentences. As a step towards building such a system, this paper presents our work on automating the classification of sentences as fair, potentially unfair, and clearly unfair. We investigate the ability of Pre-Trained Language Models (PLMs) to perform the classification by comparing Chain of Thought (CoT) prompting \cite{wei2023chainofthought} and semi-supervised fine-tuning. 

To date, prior work has primarily focused on detecting potentially unfair clauses within online Terms of Service (ToS) agreements, emphasizing fairness from a consumer-centric standpoint \cite{lippi}. In comparison, our work delves into the intricate organizational dynamics and business context inherent with respect to commercial contracts. In our knowledge, this is the first work considering fairness in commercial contracts from the perspective of non-legal stakeholders. Furthermore, our research introduces a novel dimension by acknowledging ambiguity as a potential trigger for a perception of unfairness within contractual clauses—an aspect hitherto unexplored in the literature. Finally, making the perspectives of non-legal stakeholders available to legal teams while negotiating contracts is a first attempt, and building an automated assistant for this purpose contributes to both legal and non-legal stakeholders. 

 The primary contributions of this work can be summarized as follows: (1) We empirically analyze fairness in commercial contracts from the perspective of non-legal stakeholders; (2) We introduce a novel classification task tailored specifically for commercial contracts. This task categorizes sentences into three distinct classes: fair, potentially unfair, and clearly unfair; (3) We perform a set of experiments to examine the ability of PLMs on the novel classification task. 
\vspace{-1mm}
\section{Related Work}
\vspace{-1mm}
\subsection{Defining and Categorizing Fairness}
\vspace{-1mm}
Efforts to formulate a consistent definition of fairness have extended across various domains, including law, social science, and Machine Learning (ML). Fairness in a specific context may frequently vary and become a subject of contention among individuals or groups, depending upon the social and contextual elements at play \cite{fairness-socio}. Consequently, numerous formal and informal definitions of fairness have emerged within the body of literature \cite{fairness-def1, xiang2019legal, fairness-def-2}. \citet{lippi} delineated fairness from a consumer's standpoint in online ToS agreements.

In the realm of contract negotiations, various parties and stakeholders involved are likely to have different and conflicting interests and perspectives \cite{poppo}. In such conditions, ensuring fairness becomes significant for all stakeholders involved \cite{kadefors2005fairness}. Unlike other domains, fairness in contract negotiations has broadly been categorized into procedural and distributive fairness from the perspective of stakeholders directly involved in negotiations \cite{welsh2003perceptions}. 
In our research, we encapsulate the perception of fairness among non-legal stakeholders, individuals who are not directly engaged in contract negotiations but nonetheless constitute a substantial segment of the downstream implementation process.
\vspace{-2mm}
\subsection{Detecting Unfairness in Contracts}
\vspace{-1mm}
In recent years, efforts to automate the detection of unfair contractual clauses have gained momentum. \citet{micklitz} presented an empirical argument that it is possible to partly automate the process of abstract control of fairness of clauses in online consumer contracts. A notable work for detecting unfair clauses is CLAUDETTE \cite{lippi}, a user-end tool that uses machine learning to identify and grade potentially unfair clauses in online ToS contracts. \citet{lagioia} combined learning and reasoning tasks to enable CLAUDETTE to deal with rationales for improving its performance in detecting unfair clauses and to provide legal reasons why a clause is unfair. \citet{ruggeri-a} proposed memory enhancement of Transformer models by expressing unstructured domain knowledge in natural language and evaluated their approach to unfairness detection in consumer contracts. \cite{ruggeri} used a memory attention network-based technique to explain unfairness in consumer contracts. 

Our work differs from the abovementioned literature in the following aspects: (1) While previous literature focuses on the fairness of online service agreements from a consumer perspective, this work analyzes the fairness of commercial contracts from a non-legal stakeholders’ lens. (2) Since we focus on commercial contracts, we highlight the multiple interpretations of fairness that emerge due to complex organizational structures and business processes. The existing literature does not focus on this context. (3) We uncover ambiguity as a potential factor for unfairness in commercial contracts, which, to the best of our knowledge, has not been done before. 

\vspace{-2mm}
\section{Contract Fairness from a Non-Legal Stakeholder’s Perspective}
\vspace{-2mm}
Our initial course of action was to empirically delineate the concept of unfairness in commercial contracts, particularly from a non-legal standpoint, and analyze if it differed from the viewpoint of legal stakeholders. Moreover, we sought to ascertain the extent of participation of non-legal stakeholders in contract negotiations. 

\subsection{Research Questions}

The empirical study aimed at answering the following Research Questions (RQs): \\
\textbf{RQ1}: What scenarios constitute unfairness in contracts from a non-legal stakeholder’s perspective? \\
\textbf{RQ2}: What is the impact of unfair contractual sentences on project implementation? \\
\textbf{RQ3}: How do the perceptions of unfairness among non-legal stakeholders differ from those of their legal counterparts? \\
\textbf{RQ4}: What is the extent of participation of non-legal stakeholders in contract negotiations?
\vspace{-1mm}
\subsection{Empirical Study Method}

We conducted a questionnaire-based study involving 15 non-legal stakeholders to answer the abovementioned RQs. Henceforth, we refer to these stakeholders as study participants. We used social media platforms and connections within and outside a vendor organization to select the study participants. Our selection criteria included the following factors: (a) years of experience in dealing with project implementation activities on an industrial scale, (b) exposure to implementation and delivery in diverse geographies across the globe, and (c) familiarity with contracts or regulatory text. The participants who finally volunteered for the study, denoted as P1 to P15, hail from diverse organizational backgrounds with a global customer base. Their professional roles span across the spectrum of large-scale industrial project implementation, including directors, project managers, engineers, and requirement analysts. Subsequently, we shared the findings of the study with three legal professionals, denoted as P16 to P18, who currently serve as in-house legal counsel within their respective organizations and are actively involved in contract negotiations and compliance-related activities. We provide the details about each study participant in Appendix \ref{sec:appendixA}. 

To conduct the study, we circulated a questionnaire through e-mail, inviting study participants to provide their insights based on their experience. The questionnaire consisted of two sections. Section 1 enquired about participants' backgrounds and their familiarity with contractual obligations. Queries revolved around their professional roles, years of industry experience, and prior involvement in contract negotiations. Section 2 included questions concerning the impact and scenarios of unfairness in contract text. The participants were expected to write descriptive answers for each question. Additionally, we asked participants to furnish real-world examples of unfair contract clauses that illustrated their perspectives. The questionnaire is provided in Appendix \ref{sec:appendixB}.

Two authors of this work consolidated the responses of the study participants. If the answers to any questions were unclear, the respective participants were approached again for clarification. 

\subsection{Outcomes of the Study}
\label{sec3.3}
\textit{Answering RQ1}: Each participant highlighted different scenarios of unfairness that may occur in contractual clauses. These scenarios can be grouped into four categories:

1) Imbalance in Rights or Obligations: Out of the 15 participants, 11 highlighted scenarios where there was a clear imbalance in the rights or obligations of the involved parties. For instance, 8 participants mentioned unfair cases where one party could be given unliteral rights to make decisions about the modification, termination, and jurisdiction of any agreement provisions. Five participants also explained that clauses that limited the liability of any party and imposed restrictions on claiming indemnity during the course of the project were unfair.

2) Ambiguous Language: 9 participants elucidated scenarios where the implementation boundaries were unclear, and there was a risk of non-compliance due to ambiguities in the contract. They believed that such ambiguous clauses should be perceived as unfair. For instance, P11 stated that ambiguous clauses where the imbalance between the obligations of two parties is implicitly present are potentially unfair. P8 mentioned that any clause including ambiguous language that can potentially cause conflicts between two parties and lead to loss of time and/or money should be considered unfair. 

3) Project-specific Context: 6 participants elucidated context-specific scenarios concerning timelines, cost, and other delivery-related attributes that non-legal stakeholders may consider unfair. For instance, P13 explained that a supply chain management firm may incur significant losses while delivering perishable items if the contract does not establish appropriate timelines. P13 further stated that such timelines should be agreed upon in a contract only after consulting non-legal stakeholders with relevant expertise. 

4) Human-Level Attributes: 4 participants also highlighted the importance of ensuring individual and group fairness for all stakeholders \cite{binns2019apparent}. For instance, P3 pointed out that no clause should be discriminatory or biased against any stakeholder based on social or demographic factors. 

Based on these scenarios, we formulated the following definition: 

A contractual sentence is fair if: (a) it does not cause a direct imbalance in parties’ rights or obligations, (b) it does not include ambiguous language that causes a risk of non-compliance, (c) it includes project-specific details consistent with acceptable industry standards, and (d) it does not discriminate against any stakeholder based on social or demographic factors. 

\textit{Answering RQ2}: All the study participants from a non-legal background agreed that unfair sentences directly impacted the project lifecycle, including the phases of requirements elicitation, design, and implementation. Six of them also mentioned that a few sentences impacted overall project management and raised compliance-related issues, risking legal proceedings and financial losses for the parties. 

\textit{Answering RQ3}: The three participants having a legal background agreed with the identified scenarios and definition of fairness that emerged after analyzing the responses of non-legal stakeholders. However, they did not concur with the general perception of non-legal stakeholders regarding ambiguity being a factor that directly influences the fairness of contractual clauses. They explained that contracts deliberately infuse ambiguous language to ensure coverage and flexibility. While they agreed that certain ambiguities may introduce risk, they advised against over-specifying details within the contract document to remove ambiguities. Instead, they recommended that any ambiguous details in a contract that seemed potentially unfair to non-legal stakeholders should be clarified before implementation to avoid any confusion. 

\textit{Answering RQ4}: It was evident from participants’ responses that non-legal stakeholders do not directly participate in contract negotiations. The legal stakeholders consult the heads/directors of different teams before agreeing with the terms and conditions imposed in the contract. However, these discussions are unstructured and on a high level. P12 and P14 further highlighted that reviewing contractual clauses becomes arduous for their teams due to a limited understanding of the complex language used in contracts. As a result, the general perception of people directly involved in implementation, such as requirement analysts, engineers, and project managers, is often left out.

\begin{table*}
\renewcommand{\arraystretch}{1.2}
\centering
\small
\begin{tabular}{ p{1.4cm} | p{4cm} p{3.5cm} p{5.5cm}  }
\hline
Unfairness Class & Definition & Example & Rationale for Unfairness\\
\hline
Potentially Unfair	& This category includes all contractual sentences containing ambiguous language, which can potentially cause a risk of non-compliance.	& Data Processor shall keep Personal Information sufficiently isolated from other data on the server in an appropriate manner to prevent it from being misused.	& The example is potentially unfair because what constitutes sufficient isolation and appropriate manner is vague and could be interpreted in multiple ways. This ambiguity makes it difficult for the Data Processor to comply with the obligation and makes Personal Information susceptible to misuse. \\
\hline
Clearly Unfair	& This category includes all clauses that deal with situations that lead to a direct imbalance in parties’ rights or obligations. & Purchaser may modify the policies incorporated with this Order anytime, and Supplier will be obligated to implement such modifications.	& The clause provides the Purchaser unilateral rights to modify the policies of the agreement without any prior notice to the Supplier.  \\
\hline
\end{tabular}
\caption{Description of Unfairness Classes}
\label{table:1}
\vspace{-3mm}
\end{table*}
\vspace{-2mm}
\section{Task Formulation}
\vspace{-2mm}
A key finding of our study was that the fairness perception of non-legal stakeholders might get overlooked during contract negotiations. Therefore, we decided to build an automated assistant that can leverage the reasoning capabilities of PLMs to determine unfairness in contractual sentences from the perspective of non-legal stakeholders. 

In order to build a contract negotiation assistant, we first analyzed if the unfairness scenarios highlighted by the empirical study participants could be detected automatically. While language models can be used to detect ambiguities \cite{lebanoff} and imbalances in rights and obligations \cite{lippi}, the automated detection of project-specific details necessitates knowledge of contextual features such as organizational rules, accepted industry standards, and domain expertise. These aspects can differ across projects, and therefore, cannot be generalized to build a universal negotiation assistant. Therefore, we considered unfair project-specific details out of scope for automation. Furthermore, language models are presently incapable of reasoning about human-level attributes affecting fairness \cite{fairness-socio}. Consequently, we also did not include such attributes in our classification. We plan to analyze project-specific and human-level attributes affecting the fairness of contracts in future work. 

With this background, we formulated three distinct classes for manually labeling our dataset from a non-legal stakeholder’s perspective- fair, potentially unfair, and clearly unfair. Each of these classes can be explained as follows:

\textit{Fair}: All contractual sentences that follow the definition of fairness mentioned in Section \ref{sec3.3} are labeled as fair. Additionally, sentences containing ambiguous terms that do not directly impact implementation are marked fair. For instance, the clause \textit{While visiting the Buyer's sites, Representatives of the Vendor shall conduct themselves in a businesslike manner} is vague because the term 'businesslike manner' can be interpreted by the Vendor in multiple ways. However, this clause is not unfair because it does not hold business criticality concerning the deliverables of the project. 

\textit{Potentially Unfair}: All sentences containing ambiguous language that can potentially cause a risk of non-compliance are labeled as potentially unfair. For instance, any clause dealing with Change Requests and Issue Management in the service industry is particularly susceptible to ambiguities. The accepted practice in the industry is that the vendor responsible for providing services charges extra to implement change requests. However, no extra charges can be levied by the vendor if the change request is raised due to an identified bug in the services provided. If vague terms such as ‘appropriate’ are used in clauses that contain requirements, the boundaries of the services remain ill-defined and may cause conflicts between the parties. Such conflicts can be avoided if ambiguous terms are clarified before implementation.

We note here that we do not expect the clarified details to be explicitly included in the contract drafts. As pointed out by legal stakeholders in the empirical study, contracts are written and will continue to be written in a language that is not over-specific and granular \cite{LI201798}. In this work, we focus on detecting potentially unfair sentences to clarify them between legal and non-legal stakeholders before the contracts are signed to mitigate any negative consequences. 

\textit{Clearly Unfair}: All sentences that cause a direct imbalance in parties' rights and obligations are labeled as clearly unfair. \citet{lippi} identified scenarios that can lead to an imbalance in online ToS agreements. Based on the empirical study responses, we discovered that some of these scenarios also occur in commercial contracts. These include: (1) Jurisdiction-related restrictions; (2) Choice of law governing the contract; (3) Unilateral changes to the contract; (4) Unilateral Termination of the contract; (5) Indemnity-related concerns; (6) Arbitration. 

Table \ref{table:1} provides a description of the two unfair classes, along with examples and rationales. 

\section{Dataset Creation}
\vspace{-2mm}
\subsection{Dataset Source}
\vspace{-1mm}
We used English proprietary contracts provided by a vendor organization to conduct our experiments. The dataset included 45 expired commercial contracts corresponding to nine business domains (healthcare, automotive, finance, banking, pharmaceuticals, telecom, utility, clothing-retail, and supermarket). These were digitized contracts present in HTML format. To extract text from these documents, we used HTML tags to identify the sections of each document and the contractual clauses contained within them. After identifying each contractual clause, we extracted individual sentences using a sentence tokenizer. We extracted 5940 contractual sentences from this dataset.
\vspace{-2mm}
\subsection{Labeling Process}

\begin{figure*}[!h]
\centering
\includegraphics[scale=0.85]{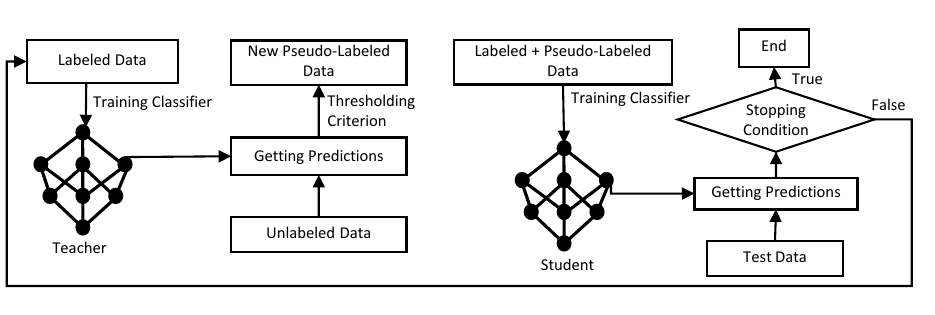}
\vspace{-3mm}
\caption{Self-Training Framework}
\label{fig1}
\vspace{-4mm}
\end{figure*}

We delegated the labeling to four non-legal stakeholders within a vendor organization (henceforth called annotators). Owing to the convoluted nature of contract text, the manual labeling of each sentence entails a substantial human effort. Therefore, we randomly selected 1200 sentences and instructed the annotators to assign one of the three identified classes to each sentence. The remaining sentences were kept unlabeled.

\begin{table}
\renewcommand{\arraystretch}{1.2}
\small
\centering
\begin{tabular}{ p{2.5cm} | c  c c c  }
\hline
 & \#1 & \#2 & \#3 & Total \\
\hline
Train before DA	& 401	& 165	& 34 & 600 \\
Train after DA	& 401	& 165	& 179 & 745 \\
Validation	& 164	& 58	& 18 & 240 \\
Test	& 235	& 99	& 26 & 360 \\
\hline
\end{tabular}
\caption{Dataset Distribution\tablefootnote{\#1 Fair; \#2 Potentially Unfair; \#3 Clearly Unfair; DA- Data Augmentation}}
\label{table:2}
\vspace{-5mm}
\end{table}

First, the annotators analyzed each contractual sentence and assigned a label denoting the appropriate class. As contracts contain vast paragraphs of text, while analyzing a particular sentence for fairness, the annotators checked whether the aspect that makes the sentence seem unfair was resolved in the preceding or succeeding text. If so, then the sentence was marked as fair. 

Every sentence underwent independent annotation by two annotators. In instances of discrepancies between the annotators, the sentence was forwarded to a third annotator tasked with assigning the final label. The annotators dedicated ten working days to completing this task, expending an average of two hours daily. Additional details about the labeling process and inter-annotator agreement scores can be found in Appendix \ref{sec:AppendixC}. The dataset distribution after labeling is provided in Table \ref{table:2}. We performed a stratified 50:20:30 split to divide the dataset into train, validation, and test sets respectively. The stratified split was based on the distribution of target classes to ensure that all training, validation, and test sets have a proportionate number of sentences for each class. We also note here that the annotators did not observe any differences while labeling contracts belonging to different domains. Therefore, the proposed unfairness categories are domain-agnostic. 
\vspace{-2mm}
\section{Experiments}
\vspace{-1mm}
We performed the ternary classification using two methods- (1) Data Augmentation and Self-Training and (2) CoT Prompting.  
\vspace{-2mm}
\subsection{Data Augmentation and Self-Training}

                    
                         
                    
                    


We encountered a significant class imbalance in the dataset, as evident in Table \ref{table:2}. We had very few sentences that were labeled as clearly unfair. Therefore, we performed data augmentation of the training set to ensure a comparable number of sentences in each class \cite{cain}. We decided to harness the generative power of ChatGPT to generate sentences corresponding to the minority class. The few-shot prompt structure used for this task was as follows: (1) We specified the system behavior that sets the context for the task. (2) We described the relevant context required by ChatGPT to generate clearly unfair sentences. (3) We provided two examples of clearly unfair sentences based on the context provided. (4) We detailed the task to be performed by ChatGPT. (5) We specified the output format. The prompt used for data augmentation is provided in Appendix \ref{sec:AppendixD}. We got the generated sentences verified by two annotators to ensure consistency with the class definition. The training dataset distribution post-augmentation is given in Table \ref{table:2}.

After augmenting the training data, we decided to use self-training \cite{self-train}, a semi-supervised learning paradigm, to leverage the large corpus of unlabeled sentences present in the dataset. The unlabeled corpus consisted of 4740 sentences.

The self-training framework we incorporated consisted of 3 main steps: (1) We trained a language model for our ternary classification task using only the labeled training data, which we refer to as the Teacher model; (2) Using the Teacher, we obtained predictions on unlabeled data and filtered them based on a confidence threshold criterion. Next, we included the unlabeled sentences that passed the criterion in the training set and treated their predictions as pseudo-labels; (3) Using the augmented dataset of labeled and pseudo-labeled sentences, we trained a new model that we refer to as the Student. We followed these steps iteratively until we reached the stopping condition. In each subsequent iteration, the Student replaced the Teacher, and we assigned pseudo-labels to the remaining unlabeled data to train a new Student. The framework is depicted in Figure \ref{fig1}. 

\begin{table*}
\renewcommand{\arraystretch}{1.2}
\small
\centering
\begin{tabular}{ p{1.3cm} | p{2.8cm}  |   c | c | c | c| c}
\hline
\textbf{Model} & \textbf{Technique} & \textbf{Fair F1} & \textbf{Potentially Unfair F1} & \textbf{Clearly Unfair F1} & \textbf{Macro F1} & \textbf{Accuracy} \\
\hline
\multirow{3}{*}{BERT} & Vanilla & 0.83	& 0.71	& 0.25 & 0.60 & 72.3\%\\
                    
                         &  Data Augmentation	& 0.82 & 0.74		& 0.51	& 0.69 & 76.7\%\\
                         
                         &  \textbf{Data Augmentation + Self-Training} & \textbf{0.85}	& \textbf{0.79} & \textbf{0.58}	& \textbf{0.74} & \textbf{84\%}\\
                        \hline
\multirow{2}{*}{Vicuna} & Direct Prompt & 0.83	& 0.61 & 0.45 & 0.63 & 63.2\% \\
                    
                         &  CoT Prompt &  0.84	& 0.73 & 0.57	& 0.71 & 75.4\% \\
                        \hline
\multirow{2}{*}{LLAMA-2} & Direct Prompt & 0.80	& 0.58	& 0.42	& 0.60 & 59.5\% \\
                    
                         &  CoT Prompt &  0.85	& 0.70 & 0.53	& 0.69 & 71\% \\
                        \hline

\end{tabular}
\caption{Results for Unfairness Detection}
\label{table:3}
\vspace{-5mm}
\end{table*}

Our thresholding criterion inputs the confidence of the model predictions and the thresholds corresponding to each class. These thresholds act as hyperparameters, and we determined them after analyzing the model confidence after each iteration. Model confidence of an individual sentence refers to the probability corresponding to the predicted class of the sentence. If the confidence exceeds the given threshold, the criterion outputs 1, and the sentence is included for augmentation. In case the output is 0, the sentence is not included. The stopping criterion was when the accuracy did not improve on the test set after consecutive iterations of self-training. This criterion is standard practice in self-training \cite{self-train}.
\vspace{-2mm}
\subsection{Chain of Thought Prompting}

Large Language Models (LLMs) have shown promising in-context learning capabilities and perform at par with fine-tuned language models on several downstream tasks \cite{liu2021pretrain}. Therefore, in addition to self-training, we incorporated Chain of Thought (CoT) prompting \cite{wei2023chainofthought} for LLMs as a multi-class classification technique. The prompting strategy is fundamentally anchored on the principle of giving the model' time to think'. This principle involves breaking down the overall task into a sequence of intermediate steps, subsequently enhancing the reasoning capabilities of LLMs beyond those offered by direct prompting.

We experimented with CoT in the few-shot setting. We also incorporated direct prompting as a baseline. The prompts used for performing the experiments are provided in Appendix \ref{sec:AppendixE}.
\vspace{-2mm}

\subsection{Experimental Setup}

In the case of self-training, we used BERT \cite{bert}, a deep bi-directional Transformer-based architecture, to build our classifier. We used the open-source version of the pre-trained bert-base-cased model available on the Hugging-Face library. We used the PyTorch library \cite{pytorch} to perform all our experiments. The details of all hyperparameters used for training the DL-based models are provided in Appendix \ref{sec:AppendixF}.

We used Vicuna (13B) \cite{vicuna2023} and LLAMA-2 (13B) \cite{touvron2023llama} as base LLMs for our prompting experiments. We selected these LLMs because they are open-source and have shown a robust performance on multiple NLP benchmarks as compared to their close-sourced counterparts such as ChatGPT. Moreover, LLAMA-2 is available under a license which makes it free for commercial usage. We used the open-source Hugging face implementation for both of these LLMs. We set the temperature parameter 0 to obtain deterministic results. The maximum context length was set to 1024 tokens. All experiments were performed using an NVIDIA V100 GPU with 60 GB RAM and 32 GB GPU memory.
\vspace{-2mm}
\section{Results and Discussion}
\vspace{-1mm}
\subsection{Classification Results}

Table \ref{table:3} shows the classwise F1 scores, macro F1 score, and the accuracy obtained on the ternary classification task using different techniques. The best-performing configuration was BERT with self-training and data augmentation, with an accuracy of 84\% on the test set. As evident from the table, using self-training and data augmentation improved the performance of BERT by a margin of 12\% as compared to its vanilla counterpart. While self-training alone improved the performance across all the classes, data augmentation especially boosted the f1 score across the clearly unfair class from 0.25 to 0.51. We provide the classwise and macro-level precision and recall scores obtained for all the experiments in Appendix \ref{sec:appendixG}.  

Self-Trained BERT outperformed few-shot CoT and direct prompting using both Vicuna and LLAMA-2. CoT prompting outperformed direct prompting for both Vicuna and LLAMA-2. CoT prompting using Vicuna performed with an accuracy of 75.4\%. We observed that the difference in accuracies of CoT using Vicuna and Self-Trained BERT ( approximately 9\%) was mainly due to the drop across potentially unfair class. Vicuna erroneously classified many potentially unfair sentences as fair. This trend can be explained because Vicuna has not been specifically fine-tuned on this task, and therefore, misses out on ambiguities contributing to potential unfairness. Vicuna marginally outperformed LLAMA-2 on all prompting experiments. We believe these emerging results on prompting LLMs are promising as they remove the overhead of manual labeling and the unavailability of data for training.
\vspace{-2mm}
\subsection{Analysis}
\vspace{-2mm}
\begin{figure}[!h]
\centering
\includegraphics[scale=0.80]{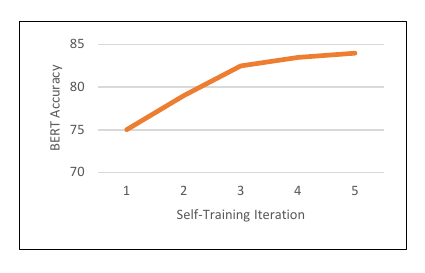}
\vspace{-3mm}
\caption{Self-Training Progression}
\label{fig2}
\vspace{-3mm}
\end{figure}

Progression of Self-Training: Figure \ref{fig2} shows the change in model performance per iteration of self-training. As self-training progresses, the overall accuracy increased from 75.2\% in iteration 1 (using only labeled data for training) to 84\% in iteration 5 on the test set. Based on how self-training progresses, we can infer that an increase in the amount of pseudo-labeled data with each iteration leads to an increase in the training set's size, enabling the model to capture features across the three classes more efficiently.

Analyzing Model Predictions: We analyzed the model predictions obtained from the test data to determine any specific patterns in sentences misclassified by Self-Trained BERT. We found that approximately 40\% of the misclassified sentences contained ambiguous terms but were followed by conditions that negate their ambiguity by establishing boundaries. For instance, the following sentence was misclassified- \textit{The system shall frequently, but in any case, at least once in 2 weeks, prompt the user to update the login password.} In this sentence, though ‘frequently’ is a vague term, it is followed by the condition, ‘once in 2 weeks’, which sets a minimum criterion for the expected frequency of changing passwords. The presence of this minimum condition eliminates any ambiguity, and therefore, it should be classified as a fair sentence. However, our model wrongly classifies it as potentially unfair. This misclassification is because cases where terms such as ‘frequently’ are followed by disambiguating phrases such as conditions and criteria rarely occur in our dataset. 

We also analyzed the reasoning generated by Vicuna for classification predictions using the CoT prompting experiments. We observed that the LLM followed the instructions and definitions in most cases. The model broke down the inference into the following steps: (a) It analyzed if the sentence is a right or obligation concerning any party. (b) It inferred if there was an imbalance in the right or obligation between parties. (c) It inferred if the sentence contained ambiguous terms. A shortcoming in the reasoning process was that the model could not correctly identify ambiguities relevant to project implementation. For instance, many sentences containing ambiguous terms such as ‘frequently’ and 'adequate' were marked fair. A similar trend was observed for LLAMA-2.

Mitigating the occurrence of misclassified sentences is critical to ensure the effectiveness of our envisioned negotiation assistant in the real world. Going forward, as future work, we plan to boost classification accuracy by incorporating human feedback into model training. A hybrid pipeline based on reinforcement learning with human feedback (RLHF) can further help LLMs discern details they tend to miss otherwise. In addition, augmenting CoT with other prompting techniques such as self-reflection also seems like a promising direction to improve results.

\vspace{-2mm}
\section{Conclusion}
\vspace{-1mm}
In this work, we analyzed and automatically detected unfairness in commercial contracts from the perspective of non-legal stakeholders. We conducted an empirical study to determine what constitutes unfairness according to non-legal stakeholders and how it differs from the perception of their legal counterparts. We discovered that non-legal stakeholders consider certain ambiguous sentences potentially unfair, even though they are legally accurate. Since these viewpoints are not directly highlighted during negotiations, we proposed building an automated negotiation assistant that can flag potentially unfair sentences according to non-legal stakeholders. As a step towards building such an assistant, we formulated a novel ternary classification task to detect unfairness in contracts. Our experiments using different PLMs and techniques revealed that self-trained BERT performed the classification with an accuracy of 84\% on a proprietary contract dataset. It outperformed CoT prompting using Vicuna by a margin of 9\%. 

In the future, we plan to build an end-to-end assistant that can reason about unfairness in contracts from the perspective of non-legal stakeholders and recommend automatically generated questions to be asked during contract negotiations. Moreover, we plan to incorporate different hybrid and prompting-driven techniques to minimize misclassification. We also plan to analyze how project-specific attributes and human-level concerns can be incorporated into the assistant to generate more personalized questions for stakeholders. 
\vspace{-2mm}
\section*{Limitations}
\vspace{-1mm}
We note a few limitations of this work: 1) We cannot share the proprietary contracts used to conduct experiments with the outside world. The ownership of this data lies with the vendor organization collaborating with us. However, to enhance the reproducibility of our work, we have released the questionnaire used to conduct the empirical study and the annotation guidelines, prompts, and hyperparameters used to conduct the experiments. The ChatGPT-generated sentences are also part of the supplementary material. 2) Since we focus on commercial contracts, we cannot ascertain the generalizability of the empirical study findings to other legal documents. 3) Although we ensure the representativeness of our dataset by incorporating commercial contracts belonging to 9 different application domains, we cannot be certain that these contracts cover all possible unfairness scenarios. 4) This work focuses on contracts written in English. We do not consider contracts written in other languages. 5) We cannot use closed-source LLMs such as ChatGPT to conduct the experiments using proprietary data. We have reported results on open-source LLMs such as Vicuna and LLAMA-2 to mitigate this limitation. 
\vspace{-2mm}
\section*{Ethics Statement}
\vspace{-1mm}
We recognize and acknowledge the potential for misuse of our proposed contract negotiation assistant, including the possibility of adversarial exploitation to deliberately introduce unfairness into contractual clauses. Additionally, there exists a risk of unintended consequences due to inaccurate predictions when dealing with unseen data, which can have a ripple effect on contract negotiations. Consequently, we stress the importance of employing this assistant judiciously and primarily as a supplementary tool in contract negotiations. It should not be considered a substitute for the critical expertise legal professionals provide. Identifying unfairness within contractual sentences is intended to serve as an aid that can enhance the efficiency of experts' reviews, thereby reducing the time and effort expended in scrutinizing contract drafts. Ensuring that the assistant is aligned with ethical practices and is coupled with expert consultation to prevent unintended ramifications is crucial.

\bibliography{anthology,custom}

\begin{thebibliography}{26}
\expandafter\ifx\csname natexlab\endcsname\relax\def\natexlab#1{#1}\fi

\bibitem[{Amini et~al.(2022)Amini, Feofanov, Pauletto, Devijver, and Maximov}]{self-train}
Massih-Reza Amini, Vasilii Feofanov, Loic Pauletto, Emilie Devijver, and Yury Maximov. 2022.
\newblock \href {https://doi.org/10.48550/ARXIV.2202.12040} {Self-training: A survey}.

\bibitem[{Binns(2019)}]{binns2019apparent}
Reuben Binns. 2019.
\newblock \href {http://arxiv.org/abs/1912.06883} {On the apparent conflict between individual and group fairness}.

\bibitem[{Chiang et~al.(2023)Chiang, Li, Lin, Sheng, Wu, Zhang, Zheng, Zhuang, Zhuang, Gonzalez, Stoica, and Xing}]{vicuna2023}
Wei-Lin Chiang, Zhuohan Li, Zi~Lin, Ying Sheng, Zhanghao Wu, Hao Zhang, Lianmin Zheng, Siyuan Zhuang, Yonghao Zhuang, Joseph~E. Gonzalez, Ion Stoica, and Eric~P. Xing. 2023.
\newblock \href {https://lmsys.org/blog/2023-03-30-vicuna/} {Vicuna: An open-source chatbot impressing gpt-4 with 90\%* chatgpt quality}.

\bibitem[{Devlin et~al.(2018)Devlin, Chang, Lee, and Toutanova}]{bert}
Jacob Devlin, Ming-Wei Chang, Kenton Lee, and Kristina Toutanova. 2018.
\newblock \href {https://doi.org/10.48550/ARXIV.1810.04805} {Bert: Pre-training of deep bidirectional transformers for language understanding}.

\bibitem[{Kadefors(2005)}]{kadefors2005fairness}
Anna Kadefors. 2005.
\newblock Fairness in interorganizational project relations: norms and strategies.
\newblock \emph{Construction Management and Economics}, 23(8):871--878.

\bibitem[{Karabulut et~al.(2020)Karabulut, Civelek, Başar, {\"O}z, and Kucukcolak}]{Karabulut2020TheRA}
Ahu~Tuğba Karabulut, Mustafa~Emre Civelek, Pınar Başar, Sabri {\"O}z, and Ali Kucukcolak. 2020.
\newblock \href {https://api.semanticscholar.org/CorpusID:229012979} {The relationships among corporate governance principles and firm performance}.
\newblock \emph{Accounting}.

\bibitem[{Lagioia et~al.(2019)Lagioia, Ruggeri, Drazewski, Lippi, Micklitz, Torroni, and Sartor}]{lagioia}
Francesca Lagioia, Federico Ruggeri, Kasper Drazewski, Marco Lippi, Hans-Wolfgang Micklitz, Paolo Torroni, and Giovanni Sartor. 2019.
\newblock \href {https://doi.org/10.3233/FAIA190305} {Deep learning for detecting and explaining unfairness in consumer contracts}.

\bibitem[{Lebanoff and Liu(2018)}]{lebanoff}
Logan Lebanoff and Fei Liu. 2018.
\newblock \href {https://doi.org/10.48550/ARXIV.1808.06219} {Automatic detection of vague words and sentences in privacy policies}.

\bibitem[{Li(2017)}]{LI201798}
Shuangling Li. 2017.
\newblock \href {https://doi.org/https://doi.org/10.1016/j.esp.2016.10.001} {A corpus-based study of vague language in legislative texts: Strategic use of vague terms}.
\newblock \emph{English for Specific Purposes}, 45:98--109.

\bibitem[{Lippi et~al.(2019)Lippi, Pa{\l}ka, Contissa, Lagioia, Micklitz, Sartor, and Torroni}]{lippi}
Marco Lippi, Przemys{\l}aw Pa{\l}ka, Giuseppe Contissa, Francesca Lagioia, Hans-Wolfgang Micklitz, Giovanni Sartor, and Paolo Torroni. 2019.
\newblock \href {https://doi.org/10.1007/s10506-019-09243-2} {{CLAUDETTE}: an automated detector of potentially unfair clauses in online terms of service}.
\newblock \emph{Artificial Intelligence and Law}, 27(2):117--139.

\bibitem[{Liu et~al.(2021)Liu, Yuan, Fu, Jiang, Hayashi, and Neubig}]{liu2021pretrain}
Pengfei Liu, Weizhe Yuan, Jinlan Fu, Zhengbao Jiang, Hiroaki Hayashi, and Graham Neubig. 2021.
\newblock \href {http://arxiv.org/abs/2107.13586} {Pre-train, prompt, and predict: A systematic survey of prompting methods in natural language processing}.

\bibitem[{Micklitz et~al.(2017)Micklitz, Pałka, and Panagis}]{micklitz}
Hans-W. Micklitz, Przemysław Pałka, and Yannis Panagis. 2017.
\newblock \href {https://doi.org/10.1007/s10603-017-9353-0} {{The Empire Strikes Back: Digital Control of Unfair Terms of Online Services}}.
\newblock \emph{Journal of Consumer Policy}, 40(3):367--388.

\bibitem[{Paszke et~al.(2019)Paszke, Gross, Massa, Lerer, Bradbury, Chanan, Killeen, Lin, Gimelshein, Antiga, Desmaison, Kopf, Yang, DeVito, Raison, Tejani, Chilamkurthy, Steiner, Fang, Bai, and Chintala}]{pytorch}
Adam Paszke, Sam Gross, Francisco Massa, Adam Lerer, James Bradbury, Gregory Chanan, Trevor Killeen, Zeming Lin, Natalia Gimelshein, Luca Antiga, Alban Desmaison, Andreas Kopf, Edward Yang, Zachary DeVito, Martin Raison, Alykhan Tejani, Sasank Chilamkurthy, Benoit Steiner, Lu~Fang, Junjie Bai, and Soumith Chintala. 2019.
\newblock \href {http://papers.neurips.cc/paper/9015-pytorch-an-imperative-style-high-performance-deep-learning-library.pdf} {Pytorch: An imperative style, high-performance deep learning library}.
\newblock In H.~Wallach, H.~Larochelle, A.~Beygelzimer, F.~d\textquotesingle Alch\'{e}-Buc, E.~Fox, and R.~Garnett, editors, \emph{Advances in Neural Information Processing Systems 32}, pages 8024--8035. Curran Associates, Inc.

\bibitem[{Pessach and Shmueli(2022)}]{fairness-def-2}
Dana Pessach and Erez Shmueli. 2022.
\newblock \href {https://doi.org/10.1145/3494672} {A review on fairness in machine learning}.
\newblock \emph{ACM Comput. Surv.}, 55(3).

\bibitem[{Poppo and Zhou(2014)}]{poppo}
Laura Poppo and Kevin~Zheng Zhou. 2014.
\newblock \href {https://doi.org/https://doi.org/10.1002/smj.2175} {Managing contracts for fairness in buyer–supplier exchanges}.
\newblock \emph{Strategic Management Journal}, 35(10):1508--1527.

\bibitem[{Ruggeri et~al.(2022)Ruggeri, Lagioia, Lippi, and Torroni}]{ruggeri}
Federico Ruggeri, Francesca Lagioia, Marco Lippi, and Paolo Torroni. 2022.
\newblock \href {https://doi.org/10.1007/s10506-021-09288-2} {Detecting and explaining unfairness in consumer contracts through memory networks}.
\newblock \emph{Artif. Intell. Law}, 30(1):59–92.

\bibitem[{Ruggeri et~al.(2021)Ruggeri, Lippi, and Torroni}]{ruggeri-a}
Federico Ruggeri, Marco Lippi, and Paolo Torroni. 2021.
\newblock \href {https://doi.org/10.48550/ARXIV.2110.00125} {Membert: Injecting unstructured knowledge into bert}.

\bibitem[{Sainani et~al.(2020)Sainani, Anish, Joshi, and Ghaisas}]{1}
Abhishek Sainani, Preethu~Rose Anish, Vivek Joshi, and Smita Ghaisas. 2020.
\newblock \href {https://doi.org/10.1109/RE48521.2020.00026} {Extracting and classifying requirements from software engineering contracts}.
\newblock In \emph{2020 IEEE 28th International Requirements Engineering Conference (RE)}, pages 147--157.

\bibitem[{Selbst et~al.(2019)Selbst, Boyd, Friedler, Venkatasubramanian, and Vertesi}]{fairness-socio}
Andrew~D. Selbst, Danah Boyd, Sorelle~A. Friedler, Suresh Venkatasubramanian, and Janet Vertesi. 2019.
\newblock \href {https://doi.org/10.1145/3287560.3287598} {Fairness and abstraction in sociotechnical systems}.
\newblock In \emph{Proceedings of the Conference on Fairness, Accountability, and Transparency}, FAT* '19, page 59–68, New York, NY, USA. Association for Computing Machinery.

\bibitem[{Singhal et~al.(2022)Singhal, Anish, Sonar, and Ghaisas}]{cain}
Anmol Singhal, Preethu~Rose Anish, Pratik Sonar, and Smita~S Ghaisas. 2022.
\newblock \href {https://doi.org/10.1145/3522664.3528604} {Data is about detail - an empirical investigation for software systems with nlp at core}.
\newblock In \emph{2022 IEEE/ACM 1st International Conference on AI Engineering – Software Engineering for AI (CAIN)}, pages 145--156.

\bibitem[{Tomlinson and Lewicki(2015)}]{contract-neg}
Edward~C Tomlinson and Roy~J Lewicki. 2015.
\newblock \href {https://doi.org/10.1177/2055563615571479} {The negotiation of contractual agreements}.
\newblock \emph{Journal of Strategic Contracting and Negotiation}, 1(1):85--98.

\bibitem[{Touvron et~al.(2023)Touvron, Martin, Stone, Albert, Almahairi, Babaei, Bashlykov, Batra, Bhargava, Bhosale, Bikel, Blecher, Ferrer, Chen, Cucurull, Esiobu, Fernandes, Fu, Fu, Fuller, Gao, Goswami, Goyal, Hartshorn, Hosseini, Hou, Inan, Kardas, Kerkez, Khabsa, Kloumann, Korenev, Koura, Lachaux, Lavril, Lee, Liskovich, Lu, Mao, Martinet, Mihaylov, Mishra, Molybog, Nie, Poulton, Reizenstein, Rungta, Saladi, Schelten, Silva, Smith, Subramanian, Tan, Tang, Taylor, Williams, Kuan, Xu, Yan, Zarov, Zhang, Fan, Kambadur, Narang, Rodriguez, Stojnic, Edunov, and Scialom}]{touvron2023llama}
Hugo Touvron, Louis Martin, Kevin Stone, Peter Albert, Amjad Almahairi, Yasmine Babaei, Nikolay Bashlykov, Soumya Batra, Prajjwal Bhargava, Shruti Bhosale, Dan Bikel, Lukas Blecher, Cristian~Canton Ferrer, Moya Chen, Guillem Cucurull, David Esiobu, Jude Fernandes, Jeremy Fu, Wenyin Fu, Brian Fuller, Cynthia Gao, Vedanuj Goswami, Naman Goyal, Anthony Hartshorn, Saghar Hosseini, Rui Hou, Hakan Inan, Marcin Kardas, Viktor Kerkez, Madian Khabsa, Isabel Kloumann, Artem Korenev, Punit~Singh Koura, Marie-Anne Lachaux, Thibaut Lavril, Jenya Lee, Diana Liskovich, Yinghai Lu, Yuning Mao, Xavier Martinet, Todor Mihaylov, Pushkar Mishra, Igor Molybog, Yixin Nie, Andrew Poulton, Jeremy Reizenstein, Rashi Rungta, Kalyan Saladi, Alan Schelten, Ruan Silva, Eric~Michael Smith, Ranjan Subramanian, Xiaoqing~Ellen Tan, Binh Tang, Ross Taylor, Adina Williams, Jian~Xiang Kuan, Puxin Xu, Zheng Yan, Iliyan Zarov, Yuchen Zhang, Angela Fan, Melanie Kambadur, Sharan Narang, Aurelien Rodriguez, Robert Stojnic, Sergey Edunov, and Thomas
  Scialom. 2023.
\newblock \href {http://arxiv.org/abs/2307.09288} {Llama 2: Open foundation and fine-tuned chat models}.

\bibitem[{Verma and Rubin(2018)}]{fairness-def1}
Sahil Verma and Julia Rubin. 2018.
\newblock \href {https://doi.org/10.1145/3194770.3194776} {Fairness definitions explained}.
\newblock In \emph{2018 IEEE/ACM International Workshop on Software Fairness (FairWare)}, pages 1--7.

\bibitem[{Wei et~al.(2023)Wei, Wang, Schuurmans, Bosma, Ichter, Xia, Chi, Le, and Zhou}]{wei2023chainofthought}
Jason Wei, Xuezhi Wang, Dale Schuurmans, Maarten Bosma, Brian Ichter, Fei Xia, Ed~Chi, Quoc Le, and Denny Zhou. 2023.
\newblock \href {http://arxiv.org/abs/2201.11903} {Chain-of-thought prompting elicits reasoning in large language models}.

\bibitem[{Welsh(2003)}]{welsh2003perceptions}
Nancy~A Welsh. 2003.
\newblock Perceptions of fairness in negotiation.
\newblock \emph{Marq. L. Rev.}, 87:753.

\bibitem[{Xiang and Raji(2019)}]{xiang2019legal}
Alice Xiang and Inioluwa~Deborah Raji. 2019.
\newblock \href {http://arxiv.org/abs/1912.00761} {On the legal compatibility of fairness definitions}.

\end{thebibliography}
\bibliographystyle{acl_natbib}

\appendix

\section{Participant Details}
\label{sec:appendixA}
We approached 40 legal and non-legal stakeholders for conducting our empirical study, out of which 18  provided concrete responses. The participation in the study was voluntary and no compensation was provided. The details corresponding to these 18 stakeholders are provided in Table \ref{table:4}.

\begin{table}[!h]
    \renewcommand{\arraystretch}{1.2}
    \centering
    \small
    \begin{tabular}{ p{0.5cm} | p{1.5cm} | p{4.5cm} }
    \hline
        ID	& Years of Experience & Current Designation / Role	\\
         \hline
         P1	& 3	& Requirement Analyst \\
        P2	& 2	& Requirement Analyst \\
        P3	& 2	& Software Engineer \\
        P4	& 4	& Software Engineer \\
        P5	& 3	& Software Engineer \\
        P6	& 6	& Solution Architect \\
        P7	& 7 & Solution Engineer \\
        P8 & 10	& Project Manager \\
        P9 & 12	& Project Manager \\
        P10	& 9	& Project Manager \\
        P11	& 15 & Senior Project Manager \\
        P12	& 20 & Business Director \\
        P13	& 24 & Delivery Head \\
        P14	& 17 & Director of Technology \\
        P15	& 16 & Director of Research \\
        P16	& 8	& Legal Counsel \\
        P17	& 6 & Legal Counsel \\
        P18	& 6	& Legal Counsel \\
         \hline
    \end{tabular}
    \caption{Details of Study Participants}
    \label{table:4}
\end{table}

\section{Questionnaire}
\label{sec:appendixB}

The questionnaire used to conduct the empirical study consisted of two primary sections. The meticulous questionnaire design aimed to obtain a thorough understanding of fairness in contracts from our study participants. 

The questions in Section 1 were as follows:
\begin{enumerate}
    \item What is your current role/designation within your organization? 
    \item How many years of experience do you have in your current line of work? 
    \item Have you previously served in other roles? Please elaborate. 
    \item Do you have experience in processing or implementing contractual obligations?
    \item Have you directly contributed to contract negotiations before?
    \item If the answer to Q5 is yes, please specify your contribution. 
    \item If the answer to Q5 is no, please explain the process of contractual governance followed in your organization. 
\end{enumerate}

The questions in Section 2 were as follows: 
\begin{enumerate}
    \item What are the issues you face when you are tasked with implementing contractual obligations?
    \item What constitutes unfairness in contractual clauses, according to you? Please specify all possible cases and provide examples to illustrate each of them.
    \item How do unfair sentences impact overall project implementation/compliance? 
    \item Do you believe non-legal stakeholders' inputs should be considered during contract negotiations? 
\end{enumerate}

\section{Dataset Annotation Details}
\label{sec:AppendixC}
The annotators incorporated for the labeling task had more than five years of experience in implementing contractual obligations. Since the labeling task was part of their daily workload, they were not compensated for it separately. We ensured that the annotators did not have conflicting interests while labeling, which could affect their judgment regarding the unfairness of clauses. We familiarized them with the task by sharing ten sentences as a sample exercise. We reviewed the ten sentences shared with each annotator and addressed any doubts and issues that came up while labeling these sample sentences. Once the annotators were familiarized with the task, they labeled 600 sentences each. 

The average inter-annotator agreement for assigning the binary labels, calculated using Cohen's kappa, was 0.64. All sentences where the two annotators differed in assigning the final label were forwarded to a third annotator. 

Some key annotation guidelines provided to the annotators were as follows:
\begin{itemize}
    \item The sentences that neither constitute rights nor obligations are fair. 
    \item If the sentence applies equally to both parties, it is fair.
    \item Sentences with redacted information should be removed from the dataset. 
    \item If the sentence applies to one party and is a right / obligation:
    \begin{itemize}
        \item If any details related to an obligation are to be decided later, after consultation with the other party, the sentence is fair. 
        \item  Any right subject to an ambiguous condition is potentially unfair. 
        \item Any right stated without any boundaries or conditions is potentially unfair. 
        \item A material obligation of a contract containing ambiguous language is potentially unfair.
        \item Any right or obligation causing a clear imbalance between two parties is clearly unfair. 
    \end{itemize}
\end{itemize}
In case of labeling potentially unfair sentences, the annotators were further asked to check if the ambiguous language used in the sentence caused a risk of non-compliance. The sentence was labeled as fair if the ambiguous language did not affect any material obligation or right concerning any party.  

\section{Data Augmentation Prompt}
\label{sec:AppendixD}
We used ChatGPT to generate sentences corresponding to the clearly unfair class in our dataset. We experimented with six prompts corresponding to the clear unfairness scenarios identified during the labeling exercise. We used the OpenAI API to interact with the model and used few-shot prompting to generate sentences that can be used for data augmentation. We generated approximately 25 sentences with each such prompt. 

One of the prompts that we used to generate clearly unfair sentences was: 

        




\noindent\fbox{%
\small

    \parbox{\linewidth}{%
        System Behavior: You are an assistant that generates contractual sentences based on the given context. \\
        
Context: A clause that allows one party to terminate the agreement unilaterally is clearly unfair. \\

Example 1: If Investor fails to cure the identified breach within thirty (30) days, Company may terminate this agreement without prior notice to Investor.\\
Example 2: Service Recipient may remove and replace any Supplier member, with or without cause, from providing the Services under this Agreement.\\

Task: Given the context and examples, generate 25 unique contractual sentences that are clearly unfair due to unilateral termination rights. \\

Output Format should be as follows: <List of Sentences>: [Sentence 1, Sentence 2, ...] \\

    }%
}
\vspace{5mm}

The other prompts used for data augmentation are provided as supplementary material to this work. We have also released the artificially generated sentences used for data augmentation as part of the package. 

\section{Classification Prompts}

\label{sec:AppendixE}
We performed the ternary classification using two prompting strategies- Direct and Chain of Thought Prompting. We experimented with both strategies in the few-shot setting by providing two examples of potentially unfair and clearly unfair classes. 
The prompts used for both techniques are as follows:

\noindent\fbox{%
\small
\renewcommand{\arraystretch}{1.4}
    \parbox{\linewidth}{%
    Direct Prompt \\
 \\
        System Behavior: You are an assistant who analyzes contractual sentences based on the given context. \\
        
Context: A contractual sentence that causes a clear imbalance in the rights or obligations between parties is clearly unfair. A sentence containing ambiguous language that causes non-compliance risk is potentially unfair. Otherwise, the sentence is fair. \\ 

Example 1: Service Recipient may remove and replace any Supplier member, with or without cause, from providing the Services under this Agreement. \\ Answer: Clearly Unfair\\

Example 2: Provider shall, upon request of Recipient, promptly enable Recipient to secure for itself, patent, trade secret, or any other proprietary rights.\\ Answer: Potentially Unfair \\

Task: Given the context and examples, classify the given contractual sentence as fair, potentially unfair, or clearly unfair. \\

Input Sentence: <input> 

    }%
}


\noindent\fbox{%
\small
    \parbox{\linewidth}{%
    CoT Prompt \\
    
        System Behavior: You are an assistant who analyzes contractual sentences based on the given context. \\
        
Context: A contractual sentence that causes a clear imbalance in the rights or obligations between parties is clearly unfair. A sentence containing ambiguous language that causes non-compliance risk is potentially unfair. Otherwise, the sentence is fair. \\ 

Example 1: Service Recipient may remove and replace any Supplier member, with or without cause, from providing the Services under this Agreement. \\ Answer: The given sentence gives termination and replacement rights to the Service Recipient. This right creates an imbalance between the Supplier and the Service Recipient because the Service Recipient can unilaterally make decisions for Supplier personnel. Therefore, the sentence is Clearly Unfair. \\

Example 2: Provider shall, upon request of Recipient, promptly enable Recipient to secure for itself, patent, trade secret, or any other proprietary rights. \\ Answer: The given sentence imposes an obligation on the Provider. However, the implementation boundaries are unclear because of the presence of vague terms such as promptly, thereby introducing a risk of non-compliance. Therefore, the sentence is potentially unfair. \\ 

Task: Given the context and examples, reason step by step and classify the given contractual sentence as fair, potentially unfair, or clearly unfair.  \\

Input Sentence: <input> 

    }%
}

\section{Hyperparameter Details}
\label{sec:AppendixF}
Initial results showed that BERT was prone to overfitting our train data. To overcome this issue, we used three strategies- (1) Early stopping based on the model accuracy on the test set; (2) L2 regularization with Adam optimizer; (3) Adding dropout to the final layer of BERT to introduce randomness in training. We used the grid-search method to obtain optimal hyperparameters for our model training. The hyperparameters used for self-training BERT are provided in Table \ref{table:5}. 

\begin{table}[!h]
\renewcommand{\arraystretch}{1.2}
\centering
\small
\begin{tabular}{  p{4cm}  p{1.8cm}   }
\hline
\textbf{Hyperparameters} & \textbf{Values} \\
\hline
Batch Size & 16 \\
Learning Rate & [1e-6, 5e-6] \\
No. of epochs & 12 \\
No. of warmup steps & 200 \\
Weight decay & 0.08 \\
Dropout & [0.2, 0.45] \\
Max sequence length & 256 \\
Confidence Threshold  Range & [0.6, 0.85] \\
\hline
\end{tabular}
\caption{Hyperparameter Details}
\label{table:5}
\end{table}

\begin{table*}
\small
\centering
\begin{tabular}{ p{1.2cm} | p{2.4cm}  |  c c c | c c c | c c c | c c c}
\hline
\textbf{Model} & \textbf{Technique} & \multicolumn{3}{|c|}{\textbf{Fair}} & \multicolumn{3}{|c|}{\textbf{Potentially Unfair}} & \multicolumn{3}{|c|}{\textbf{Clearly Unfair}} & \multicolumn{3}{|c}{\textbf{ Macro}}\\
\hline
& & P	& R	& F1 & P & R & F1 & P & R & F1 & P & R & F1\\
\hline
\multirow{3}{*}{BERT} & Vanilla & 0.82	& 0.83	& 0.83	& 0.74	& 0.68	& 0.71	& 0.24	& 0.26	& 0.25 & 0.60 & 0.59 & 0.60\\
                    
                         &  Data Augmentation &  0.82 & 0.82	& 0.82	& 0.78	& 0.70	& 0.74	& 0.52	& 0.49	& 0.51	& 0.71 & 0.67 & 0.69\\
                         
                         &  \textbf{Data Augmentation + Self-Training} & \textbf{0.88}	& 0.84	& \textbf{0.85}	& \textbf{0.80}	& \textbf{0.79}	& \textbf{0.79} &	\textbf{0.60}	& 0.56	& \textbf{0.58}	& \textbf{0.76} & \textbf{0.73} & \textbf{0.74}\\
                        \hline
\multirow{2}{*}{Vicuna} & Direct Prompt & 0.81 & 0.84	& 0.83	& 0.62	& 0.60	& 0.61	& 0.47	& 0.43	& 0.45	& 0.63 & 0.62 & 0.63 \\
                    
                         &  CoT Prompt &  0.83	& \textbf{0.86}	& 0.84	& 0.77	& 0.70	& 0.73 &	0.57 &	\textbf{0.57}	& 0.57	& 0.72 & 0.71 & 0.71 \\
                        \hline
\multirow{2}{*}{LLAMA-2} & Direct Prompt & 0.79	& 0.82	& 0.80	& 0.59	& 0.54	& 0.58	& 0.40	& 0.44	& 0.42	& 0.59 & 0.60 & 0.60 \\
                    
                         &  CoT Prompt &  0.86	& 0.85	& 0.85	& 0.73	& 0.68	& 0.70 &	0.56	& 0.50	& 0.53	& 0.72 & 0.68 & 0.69 \\
                        \hline

\end{tabular}
\caption{Precision and Recall Scores for Unfairness Detection}
\label{table:6}
\vspace{-5mm}
\end{table*}

\section{Extended Results}
\label{sec:appendixG}
Table \ref{table:6} shows the classwise precision, recall, and F1 scores obtained for different experiments conducted in this work.

\end{document}